\documentclass[times,twocolumn,final]{elsarticle}%正确
\usepackage{amssymb} 
\usepackage{times}
\usepackage{epsfig}
\usepackage{graphicx}
\usepackage{amsmath}
\usepackage{subfigure}
\usepackage{float}
\usepackage{multirow}
\usepackage{amsmath,amssymb}
\usepackage{booktabs}
\usepackage{hyperref}
\usepackage{booktabs}
\journal{Journal of \LaTeX\ Templates}
\usepackage{pgfplots} 
\usepackage{bbding}
\usepackage{soul}
\usepackage{amsmath}
\usepackage{makecell}
\usepackage[misc]{ifsym}
\usepackage{prletters}
\usepackage{framed,multirow}

\usepackage{color}

\begin{document}

\begin{frontmatter}

\title{Improving 2D face recognition via fine-level facial depth generation and  RGB-D complementary feature learning}
%%\title{An Effective Gait Recognition Based on Joint Knowledge Distillation}

\author[1]{Wenhao Hu} 
\ead{wenhao.hu@sdust.edu.cn}

%\author[2]{Peng Zhang}
%\ead{pengzhang_skd@sdust.edu.cn}

%\author[3]{Qi Li}
%\ead{qi.li@ia.ac.cn}

%\author[1]{Caifeng Shan\corref{cor1}} 
%\cortext[cor1]{Corresponding author.} 
%\ead{caifeng.shan@sdust.edu.cn}

\address[1]{College of Electrical Engineering and Automation, Shandong University of Science and Technology, Qingdao 266590, China}
%\address[2]{College of Computer Science and Engineering, Shandong University of Science and %Technology, Qingdao 266590, China}
%\address[3]{Center for Research on Intelligent Perception and Computing (CRIPAC), National Laboratory %of Pattern Recognition (NLPR), Beijing 100190, China}

%% abstract
\begin{abstract}
Face recognition in complex scenes suffers severe challenges coming from perturbations such as pose deformation, ill illumination, partial occlusion. Some methods utilize depth estimation to obtain depth corresponding to RGB to improve the accuracy of face recognition. However, the depth generated by them suffer from image blur, which introduces noise in subsequent RGB-D face recognition tasks. In addition, existing RGB-D face recognition methods are unable to fully extract complementary features. In this paper, we propose a fine-grained facial depth generation network and an improved multimodal complementary feature learning network. Extensive experiments on the Lock3DFace dataset and the IIIT-D dataset show that the proposed FFDGNet and I MCFLNet can improve the accuracy of RGB-D face recognition while achieving the state-of-the-art performance.

\end{abstract}

%% keyword
\begin{keyword}
%%\KWD 
Depth generation  \sep  RGB-D face recognition \sep multi-modality fusion
\end{keyword}

\end{frontmatter}

%%in
\section{Introduction}
\label{sec1}

Face recognition (FR) in complex scenes achieves great improvements, but it remains challenging due to  large illumination variation, pose deformation, facial expression and partial occlusion. To address these issues, some methods \cite{cui2018improving, pini2018learning, arslan2019face, baby2020face, chiu2021high} resort to depth estimation to provide compensatory information, so that improving the accuracy of FR. However, these methods typically use $l_1$ and $l_2$ loss to reduce the overall pixel difference between the generated depth and the ground truth (GT) depth, which lead to ill-posed problem in depth estimation. The ill-posed problem may decrease robustness of the depth estimation model, and potentially lead to facial blur on the generated depth images. In addition, the approachs \cite{cui2018improving,lee2016accurate,zhang2018rgb}  of RGB-D FR also affect the contribution of the generated depth images. Existing RGB-D recognition methods do not impose appropriate constraints on the modality-specific feature extraction branches, which prevents them from fully learning complementary features in multi-modal inputs, thus affecting the performance of RGB-D face recognition.

\begin{figure}[t]
\begin{center}
   \includegraphics[width=0.9\linewidth]{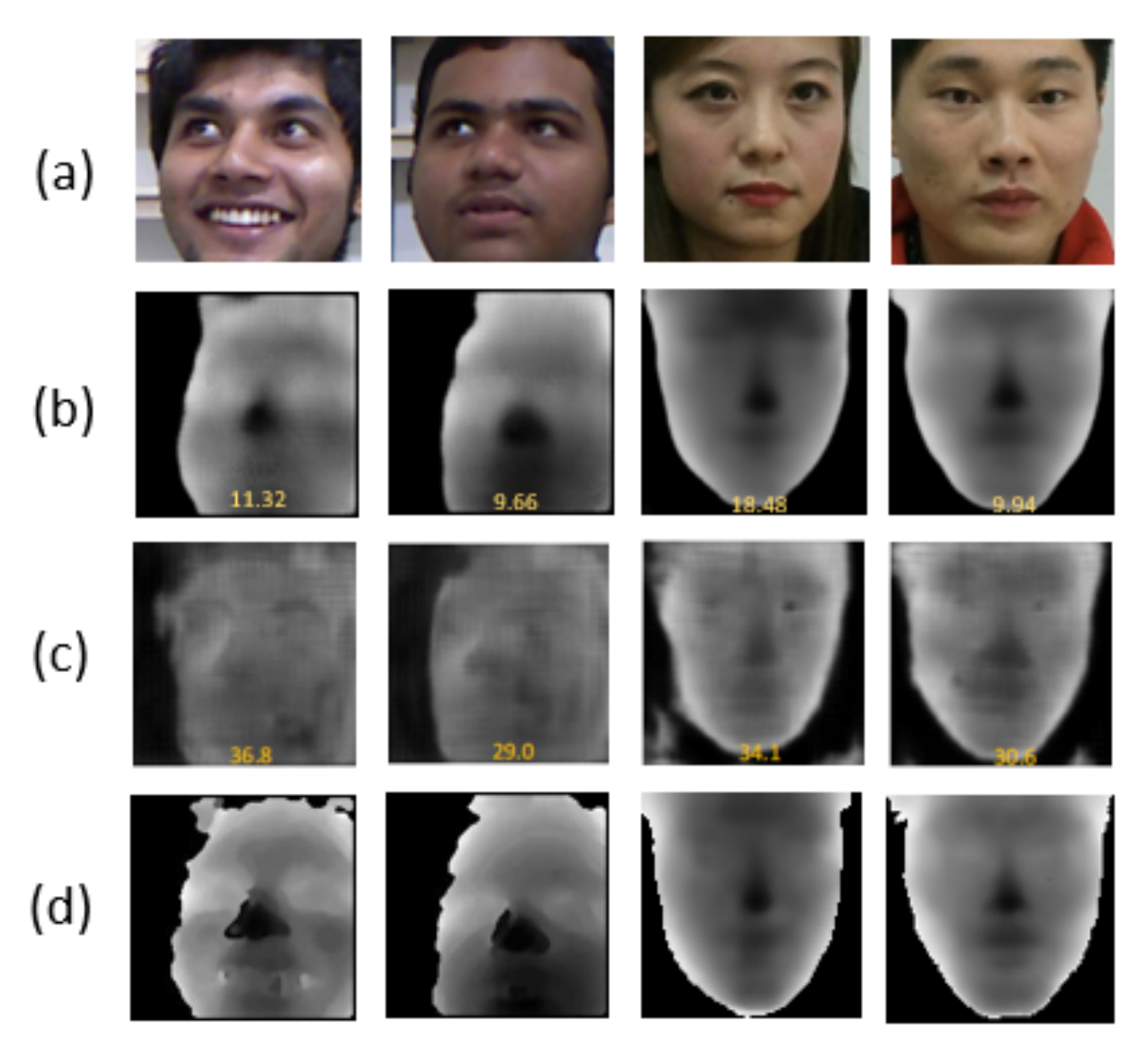}
\end{center}
   \setlength{\abovecaptionskip}{-0.2cm}\caption{Comparison of facial depth images generated by our method and baseline \cite{cui2018improving}, the value on depth image is the MAE between the generated depth image and its ground-truth (GT) depth image. (a) input RGB images, (b) the generated depth images and MAE from our method, (c) the generated depth images and MAE from baseline \cite{cui2018improving}, (d) the GT depth images.}
\label{figure_1}
\vspace{-1.0em}
\end{figure}

\begin{figure*}[t]

	\begin{center}
		\includegraphics[width=0.9\linewidth]{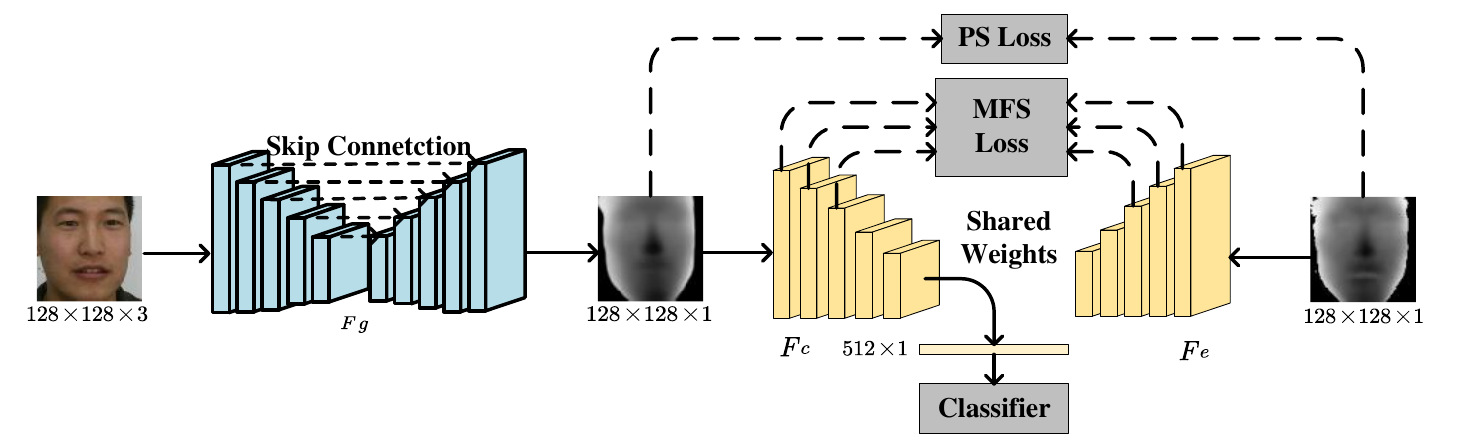}
	\end{center}
	\setlength{\abovecaptionskip}{-0.2cm}\caption{Architecture of the proposed FFDGNet. UNet generates a corresponding depth map from the RGB input, and then the generated depth map and the GT depth map are separately fed into the classification network and feature extraction network. The PS loss is used to reduce the average pixel difference between them, while the MFS loss is used to constrain the similarity between shallow feature maps, helping UNet learn more low-level features to improve the visual quality of the generated depth map. The classifier is used to ensure the discriminability of the generated depth map.}
	\label{figure_2}
\end{figure*}

To address the above issues, we introduce a novel two-stage framework that includes a fine-grained facial depth generation network (FFDGNet) and an improved multi-modality complementary feature learning network (IMCFLNet). In FFDGNet, we feed the generated depth and corresponding GT depth into a classification network and a feature extraction network with shared weights, respectively. Considering that the features from shallow layers of the convolutional neural networks (CNNs) are low-level features such as contours, we introduce a multi-level feature loss to minimize the distance between the feature maps output by the two networks at shallow layers, thereby maximizing the similarity between the generated depth images and GT depth images on low-level features, and allowing the generated depth images to contain richer facial details. In IMCFLNet, we cascade a feature separation module behind two independent feature extraction branches. This module employs four mapping functions to separate modality-specific features and modality-shared features from the feature representations of two modalities extracted by feature extraction branches. To extract complementary features, we introduce a cross-modality identity consistency loss (CIC loss) to maximize the similarity between modality-shared features, and a cross-modality feature exclusion loss (CFE loss) to minimize the similarity between modality-specific features. Extensive experiments performed on the Lock3DFace and IIIT-D dataset show that the proposed FFDGNet could generate fine-grained depth images, as illustrated in Fig. \ref{figure_1} and Table \ref{table_1}. In addition, compared to the baseline \cite{cui2018improving}, RGB-D FR accuracy of the generated depth by FFDGNet on MCFLNet achieve 0.8\% and 2.55\% improvement on two datasets, respectively. Ablation studies demonstrate the effectiveness of the proposed FFDGNet and MCFLNet.

In summary, our contribution is three-fold,
\begin{itemize}
	\setlength{\itemsep}{0pt}
	\setlength{\parsep}{0pt}
	\setlength{\parskip}{0pt}
	\item A fine-grained facial depth generation network is designed to generate depth images containing more facial details for RGB-D FR.
	\item A multi-modality complementary feature learning network is designed to learn more complementary features from RGB-D inputs.
	\item Extensive experiments performed on two public datasetsdemonstrate the effectiveness of the proposed model.
\end{itemize}

\begin{figure*}[htbp]

	\begin{center}
		\includegraphics[width=0.9\linewidth]{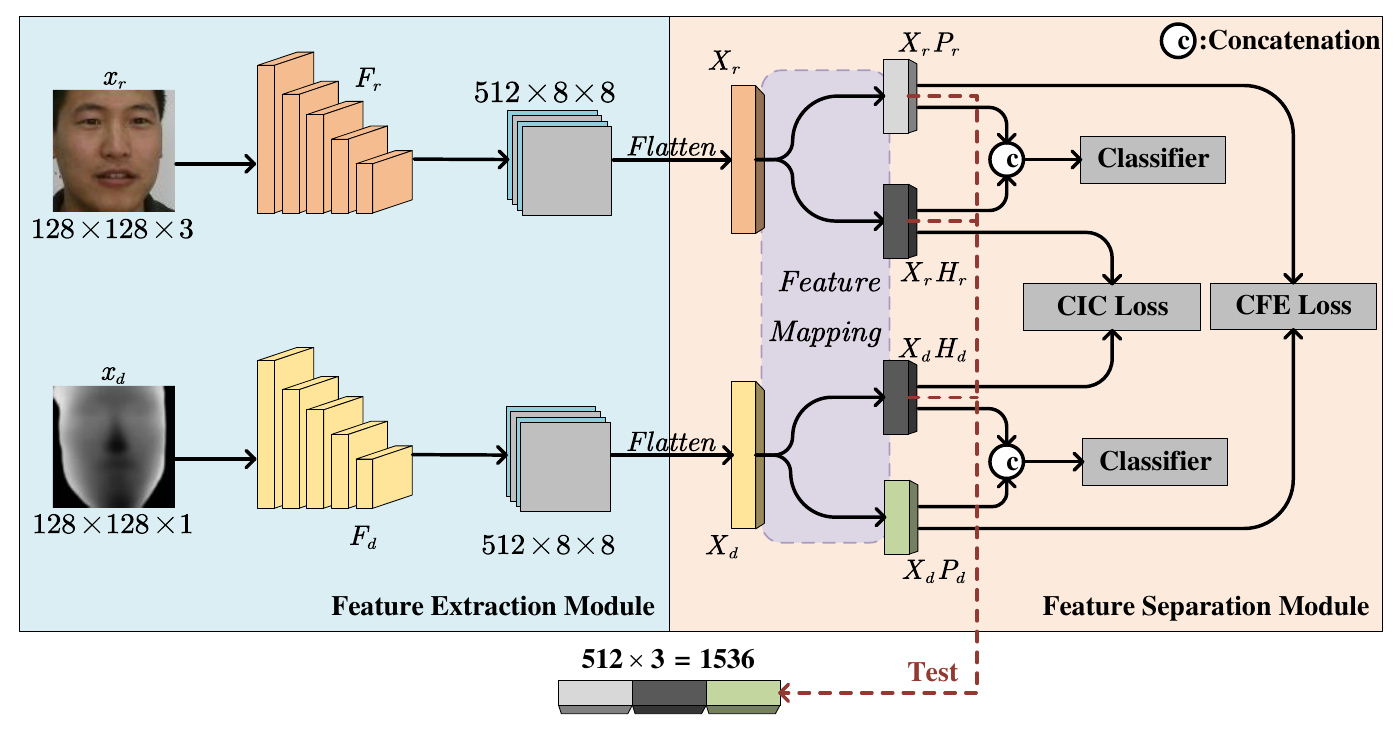}
	\end{center}
	\setlength{\abovecaptionskip}{-0.2cm}\caption{Architecture of the proposed MCFLNet. The feature extraction module first extracts the feature representations of the input RGB-D pair, and then the feature separation module introduces two loss functions to minimize the similarity between the modality-shared features and  maximize the similarity between modality-specific features to constraining the network to learn more complementary features.} 
	\label{figure_3}

\end{figure*}

\section{Related work}

\subsection{ Facial depth generation} 
The methods of facial depth generation can be categorized into three main approaches: prior-model-based, landmark-based, and encoder-decoder-based methods. Prior-model-based methods typically utilize or generate a prior model as auxiliary information during the generation process. For example, Sun et al. \cite{sun2011depth} formulate the rotation and translation of face images from frontal to non-frontal as a constrained independent component analysis model and use the CANDIDE 3D face model to transform the over-complete ICA into a general ICA problem. Aissaoui et al. \cite{aissaoui2014rapid} use the ASM algorithm to obtain a parallax model of the face, which is then used as the prior model for depth estimation. However, since the statistical-based regression used in prior-model-based methods can be inaccurate, these approaches may result in loss of some facial details during depth generation.

Landmark-based methods typically rely on the established mapping relationship between 2D feature points and 3D feature points to achieve depth estimation. Yang et al. \cite{yang2018research} formulated the pose, shape, and depth estimation problem as a nonlinear least-squares model, where the depth value of feature points in the 2D face image is obtained by minimizing the distance between the feature points of the 2D face image and the corresponding projection points on the 3D model. Xiao et al. \cite{xiao20143d} applied the coupling dictionary learning method based on sparse representation to learn the underlying mapping between 2D and 3D feature points. However, landmark-based methods can only ensure the general shape of the generated depth face and may not capture all the facial details. 
 
Encoder-decoder-based methods usually take RGB image as input and perform depth generation by constraining the Euclidean distance between the output of the decoder and the target image. For example, Baby et al. \cite{baby2020face} and Reesink et al. \cite{reesink2020creating} use UNet \cite{ronneberger2015u} as the generation model and $l_1$ is used to reduce the pixel difference between the generated depth image and the GT depth image. The generated image has a realistic appearance but insufficient discrimination for FR. Cui et al. \cite{cui2018improving} designed a cascade network of FCN and CNN, which added a discrimination constraint to improve the discrimination of the generated depth images. However, the depth images generated by this method still have some facial blur and contour distortion, which would introduce noise in subsequent RGB-D FR. Therefore, we designed a fine-grained facial depth generation network to eliminate facial blur and generate depth images containing richer facial details.

\subsection{RGB-D face recognition} 
Existing RGB-D FR methods aim to fuse RGB and depth features effectively. Based on the fused features, these methods can be categorized into image-level fusion and feature-level fusion. Image-level fusion involves the fusion of raw images or feature maps of RGB and depth modalities for FR. For instance, Uppal et al. \cite{uppal2021two} concatenate raw RGB and depth images in the channel dimension before feeding them to the recognition network. They then use LSTM and max-average pool to learn two-level attention for FR. On the other hand, Xiao et al. \cite{xiao2020application} use Gabor transform to obtain feature maps of directional properties and spatial density, which they concatenate as the input data of the DCNN to obtain more salient features. However, without considering the feature differences between two modalities, the single-branch feature extraction network is unable to fully extract their respective discriminative features.

Feature-level fusion, on the other hand, involves the separate extraction of features from different modalities, followed by the fusion of feature representations of RGB and depth face images using concatenation or element-wise operations. Cui et al. \cite{cui2018improving} concatenate two 1024-dimensional feature vectors to obtain a 2048-dimensional feature vector for RGB-D FR, while Cui et al. \cite{cui2018rgb} explore the effect of element-wise addition and average for RGB-D FR accuracy. 
However, without mutual constraints, two independent feature extraction branches may suffer from complementary feature extraction deficiency. Therefre, we designed a multi-modality complementary feature learning network to extract more complementary features for RGB-D FR.

%%-------------------------------------------method-------------------------------------------------
\section{Method} 

\subsection{Fine-grained facial depth generation}  
\label{Fine}
To  alleviate image blur on generated depth images, we propose the FFDGNet as shown in \ref{figure_2}.
FFDGNet consists of three modules: depth generation, feature refinement, and identity preservation module, where feature refinement module includes pixel-level constraint and feature-level constraint. In the deep generation module, we employ UNet \cite{ronneberger2015u} to realize the conversion from RGB to depth images. Then the generated depth image is fed into a classification network based on resnet-18 \cite{he2016deep} for identity preservation. In order to raise the similarity between the generated depth and the GT depth, we utilize the pixel-level constraints to minimize the average pixel difference of them following the baseline \cite{cui2018improving}. However, this may lead to blur on the generated depth image. To alleviate the issue, we introduce the feature-level constraint. Specifically, we first introduce a feature extraction network that is also based on resnet-18 and shares the weight parameters of the convolutional layers with the classification network. Then we input the generated depth and the corresponding GT depth into the classification network and the feature extraction network, respectively. Considering that the features from the shallow layers of CNNs are low-level features such as contours, we first extract the feature maps output by the first $n$ convolutional layers of both networks. Then, we introduce multi-level feature similarity loss (MFS loss) to minimize the pixel difference between the corresponding feature maps of two networks to help the UNet learn more low-level features, thus generating depth images that contain more facial details.

For a given RGB input $X$, we synthesize its corresponding facial depth $Y'$ using a generated model $F_g$. To minimize the distance between $Y'$ and the corresponding GT depth $Y$, a pixel similarity (PS) loss is defined as follows:
\begin{equation}
	L_{PS} = |Y'-Y|,	
\end{equation}
where $Y'=F_g(X,W_g)$, $W_g$ are parameters of the generated model $F_g$. To generate depth images with richer facial details, we respectively send the generated depth  and GT depth into the classification network and feature extraction network, then we minimize the distance between the corresponding outputs of the first $n$ corresponding convolution layers of the two networks. The multi-level feature similarity (MFS) is defined as follows:
\begin{equation}
	L_{MFS}=\frac{1}{n}\sum_{l=1}^n|F_c^l(Y',W_c^l)-F_e^l(Y,W_e^l)|,
\end{equation}
Where $F_c$ denotes the classification network and $F_e$ denotes the feature extraction network, $W_c$ and $W_e$ are their parameters. $F_c^l$ and $F_e^l$ denotes the $l$-th convolution layer of $F_c$ and $F_e$, respectively. In our work, $n$ is empirically set to 3. To ensure the discrimination of the generated facial depth images, we contrain the network with an identity loss as follows:
\begin{equation}
	L_{dis} = \frac{1}{m}\sum_{i=1}^m softmax(y_i,F_c(Y_i',W_c)),
\end{equation}
where $m$ is the number of face images in the training set. $y_i$ denotes the label of  input image, $Y_i'$ is the generated depth image. During training, we simultaneously optimize all the models. Thus, the overall loss of the proposed FFDGNet is 
\begin{equation}
	L_{Total} =\lambda_1L_{PS}+\lambda_2L_{MFS}+L_{dis},
\end{equation}
where $\lambda_1$ and $\lambda_2$  are the two hyper-parameters used to balance the three loss terms. Inspired by Liang et al. \cite{liang2023mask}, we set $\lambda_1$ and $\lambda_2$ as $L_{PS}/L_{Total}$ and $L_{MFS}/L_{Total}$, respectively. The purpose is to let the network dynamically adjust the weight according to the learning process of the task, and accelerate the convergence of the FFDGNet.

\subsection{Multi-modality complementary feature learning network}
\label{Complementary}
To learn complementary information across two modalities, we introduce the MCFLNet. As shown in Fig. \ref{figure_3} . MCFLNet consists of three modules: feature extraction, feature separation and identity preservation module. In the feature extraction module, we use two independent resnet-18 networks to extract feature representations of two modalities. However, without mutual constraint, these two branches cannot fully extract complementary features. Therefore, we cascade a feature separation module after the feature extraction module. Specifically, we first flatten the feature representation of 512x8x8 dimensions into the feature embedding of 32768x1 dimensions. Then, we use four FC layers to map the feature embeddings of two modalities into four feature subspaces: RGB-specific, RGB-shared, depth-shared, and depth-specific, with the feature dimension of $512\times1$. As inputed RGB-D images have the same identity features and similar low-level features, we introduce a cross-modal identity consistency loss (CIC loss) that minimizes the element-wise differences between the RGB-shared feature embedding and the depth-shared feature embedding. To extract more complementary features, we introduce a cross-modal feature exclusion loss (CFE loss) that maximizes the element-wise differences between the RGB-specific feature embedding and depth-specific feature embedding. In the identity preservation module, we feed above four feature embedding into two classifier to ensure the discriminability of them.

Specifically, for the given RGB-D inputs $x_i$, where $i$ $\in$ \{r = RGB, d = depth\}, $X_i$ is the Flattened feature embeddings. Then, we use four FC layers $P_i$ and $H_i$ to learn modality-specific features and modality-shared features, respectively. To increase the similarity between modality-shared feature, the cross-modal identity consistency (CIC) loss is defined, i.e.,
\begin{equation}
	L_{CIC} = |H_r(X_r, W_r^H)-H_d(X_d,W_d^H)|,
\end{equation}
where $X_i = Flatten(F_i(x_i,W_i))$, $F_i$ is the feature extraction network of modality $i$, and $W_i$ is its parameters. $W_i^H$ is the parameters of $H_i$ which extract the modality-shared features from the modality $i$. To reduce the similarity between modality-specific, the cross-modal feature exclusion (CFE) loss is defined, i.e.,
\begin{equation}
	L_{CFE} = normalize(\frac{<X_r^{sp},X_d^{sp}>}{||X_r^{sp}||_2||X_d^{sp}||_2}),
\end{equation}
where $X_i^{sp}=P_i(X_i, W_i^P)$, $W_i^P$ is the parameters of $P_i$ which extract modality-specific features from the modality $i$. $normalize(C_i)=\frac{1}{b}\sum_{i=0}^bC_i\times0.5+0.5$, $b$ is the batch size and $C_i$ is the cosine similarity between the modality-specific features of the $i-th$ pair in a batch. In addition, to ensure the discrimination of modality-specific features and modality-shared features, we contrain the network with an identity loss as follows:
\begin{equation}
	\begin{aligned}
		L_{dis} = \frac{1}{2}\sum_{i\in\{r,d\}}\sum_{j=1}^m arcsoftmax(y_j, FC(cat(X_i^{sp},X_i^{sh}))),
	\end{aligned}
\end{equation}
where $y_j$ denotes the label of input RGB-D pair, $m$ is the number of face images in training set, $cat(.,.)$ denotes concatenation operation along channel direction, $FC$ is the classification layer. During training, we minimize $L_{CIC}$ loss and $L_{CFE}$ loss to constrain the network to learn modality-specific features and modality-shared features for more discriminative features. Meanwhile, $L_{dis}$ is used to ensure the above feature embeddings are discriminative. Thus, the overall loss of the proposed MCFLNet is defined as follows:
\begin{equation}
	L_{Total} = L_{CIC}+L_{CFE}+\lambda L_{dis},
\end{equation}
where $\lambda$ is the trade-off hyper-parameter that controls the importance of feature discriminability. In this paper, $\lambda$ is empirically set to 0.5.

\section{Experiments}
\label{sec4}
In this section, we evaluate our proposed method on two datasets, i.e., Lock3DFace and IIIT-D. It should be noted that all recognition networks are firstly pre-trained following \cite{cui2018improving}, and then they are fine-tuned on Lock3DFace and IIIT-D. The final testing features are obtained by the concatenation of modality-shared features and modality-specific features.
\subsection{Datasets} 
\textbf{Lock3DFace} \cite{zhang2016lock3dface} contains 5711 video sequences of 509 subjects, which with variations in expression, occlusion, pose and time. This dataset is divided into S1 and S2 sessions, where the S2 session was captured by 169 subjects from S1 half a year later. In this paper, We extract 33,837 RGB-D images from all the video sequences and randomly select 340 subjects (22,686 images) as the training set, the remaining 169 subjects (11,151 images) are used for testing according to the \cite{cui2018improving}. 

\textbf{IIIT-D} \cite{goswami2013rgb} contains 4603 RGB-D images of 106 subjects, including variation in pose and expression. In this paper, we randomly selected 72 subjects (3,521 images) as training set, and the remaining 32 subjects (937 images) are used for testing followed by  \cite{cui2018improving}.
\subsection{Implementation details}
To carry out the experiments, we adopt the same pre-training strategy as \cite{cui2018improving}. During training, the batch-size of FFDGNet and MCFLNet is set to 32 and 4, respectively. Both networks use SGD optimizer and their initial learning rates are set to 0.01 and 0.001, respectively. We decay the learning rate if the metric of the validation set is not better than the current best value after 5 consecutive epochs. In practice, the decay rate is set to 0.5. The proposed methods are implemented with PyTorch and run on NVIDIA Quadro RTX8000 GPU.

During testing, we follow the setting in \cite{cui2018improving}. We select a frontal neutral image of each subject from the S1 session as the gallery set, and the rest images as the probe set for Lock3DFace. For IIIT-D, we select a near-frontal neutral image  as the gallery set and other images as the probe set.
\subsection{Metrics}
\textbf{Depth generation}. Following the previous work \cite{cui2018improving,liu2018joint}, we adopt mean absolute error (MAE) to evaluate the quality of generated face depth. Denoting $y$ is the generated depth , $Y$ is the GT depth, the MAE is defined as $MAE= \frac{1}{N}\sum_{i=1}^N|y-Y|$, where $N$ is the total pixel numbers of $Y$ and i denote $i$-th pixel.

\textbf{Face recognition}. Rank-1 identification accuracy is a commonly used metric for FR \cite{cui2018improving,zhang2018rgb,chiu2021high}. In this paper, each face image in the probe set is matched with face images in the gallery set one by one using cosine similarity measure. The rank-1 identification accuracy is defined as
\begin{equation}
	p=\frac{N_{correct}}{N_{total}}\times100\%,
\end{equation}
where $p$ is the identification accuracy, $N_{correct}$ is the number of all correctly classified samples, $N_{total}$ is the all number of samples in probe set.

\begin{figure}[t]
\vspace{-2.0em}
	\begin{center}
		\includegraphics[width=0.9\linewidth]{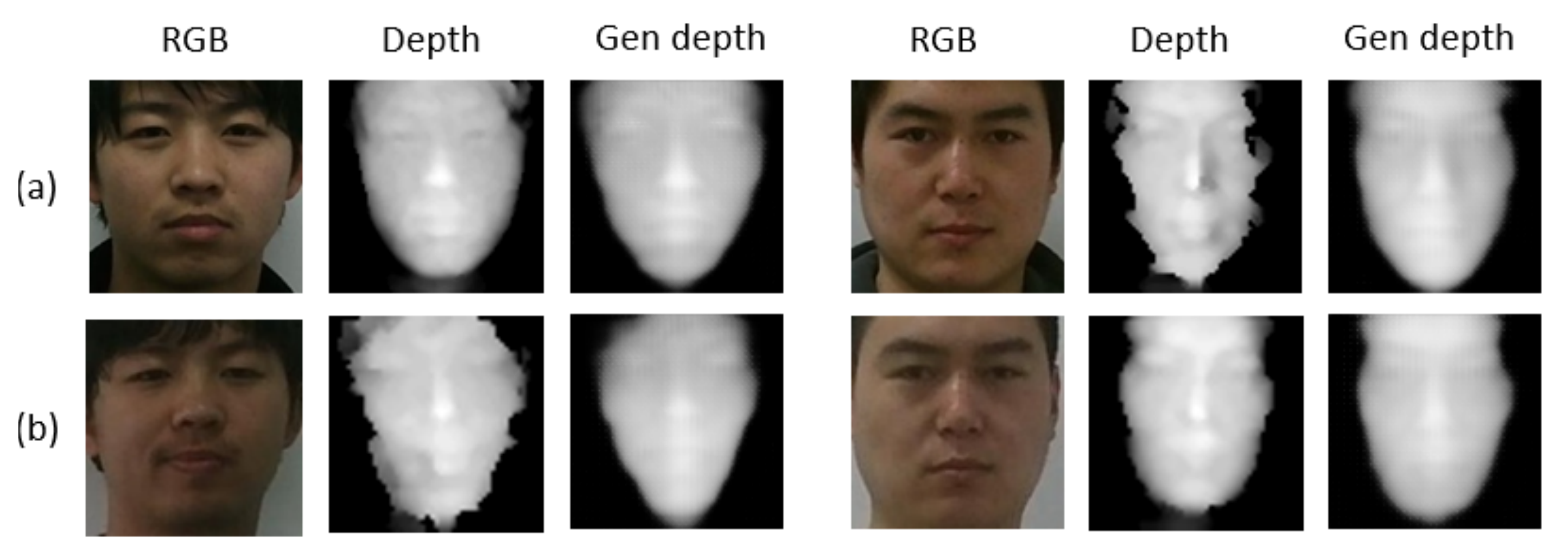}
	\end{center}
	\setlength{\abovecaptionskip}{-0.2cm}\caption{Examples of RGB and depth images of the same subject in s1 and s2 session. (a) A subject in s1 session, (b) The corresponding subject in s2 session }
	\label{figure_4}
	\vspace{-1.0em}
\end{figure}

\subsection{Results}
\textit{Effectiveness of FFDGNet}. To evaluate the effectiveness of FFDGNet, we first compared  the generated facial depth images from FFDGNet and the baseline \cite{cui2018improving}. As shown in Fig. \ref{figure_1}, our facial depth images exhibit finer contours and higher similarity with the ground truth depth images. Additionally, we conducted a quantitative analysis by calculating the MAE of the depth images generated by FFDGNet and the baseline. Table \ref{table_1} shows that our method reduces the MAE from 22.70 to 18.71 on Lock3DFace and from 28.20 to 22.80 on IIIT-D, respectively. This confirms our above statement.

In our research, we aim to enhance the RGB-D FR performance by generating depth images. Therefore, we trained two feature extraction networks based on resnet-18 using RGB and depth images generated by FFDGNet and the baseline. The test results are shown in Table \ref{table_2}, which indicate that the depth images generated by our method have better RGB-D FR performance. Notably, in the TM subset of Lock3DFace, the accuracy is improved by about 3.3\%. This may be attributed to the fact that images in the TM subset and other subsets are captured under different time conditions. As shown in Fig. \ref{figure_4}, the GT depth images of the same person captured in s1 session and s2 session are different, but the depth image generated by our FFDGNet reduces this difference. Additionally, we separately calculated the average pixel difference of the generated depth and the GT depth in s1 and s2, i.e., the MAE between the GT depths is 42.37, and the MAE between the generated depths is 34.77, which confirms the previous statement.

\begin{table}[t]

\begin{center}
\setlength{\belowcaptionskip}{0.4cm}
\caption{Comparison of MAE for face depth images generated by FFDGNet and baseline on Lock3DFace and IIIT-D.}
\setlength{\tabcolsep}{7mm}{
\begin{tabular}{ccc} 
\hline
    \multicolumn{1}{c}{\multirow{2}{*}{Dataset}} & \multicolumn{2}{c}{pixel-wise MAE$\downarrow$}  \\ \cline{2-3}
    & Ours & [1] \\ \hline
    \multicolumn{1}{c}{Lock3DFace} & 18.71 & 22.70 \\ 
    \multicolumn{1}{c}{IIIT-D} & 22.80 & 28.20 \\
\hline 
\end{tabular}}
\label{table_1}
\end{center}
\vspace{-1.0em}
\end{table}

\begin{table}[htbp]
\vspace{-1.0em}
	\begin{center}
		\setlength{\belowcaptionskip}{0.4cm}
		\caption{The rank-1 identification accuracy of RGB and face depth images generated by baseline and FFDGNet. The feature extraction network is based on ResNet18. NU: neutral, FE: expression, OC: occlusion, PS: pose, TM: time, AVG: average. }
		\setlength{\tabcolsep}{1mm}{
			\begin{tabular}{cccccccc} 
				\hline
				\multicolumn{1}{c}{\multirow{2}{*}{Input}} & \multicolumn{6}{c}{Lock3DFace} & IIIT-D \\ \cline{2-7}
				\multicolumn{1}{c}{\multirow{2}{*}{}} & NU & FE & OC & PS & TM & AVG & AVG \\ \hline 
				\multicolumn{1}{c}{\multirow{1}{*}{RGB}} & 100 & 99.74 & 100 & 98.75 & 81.09 & 95.54 & 96.46 \\
				\multicolumn{1}{c}{\multirow{1}{*}{RGB+D \cite{cui2018improving}}} & 100 & 99.59 & 100 & 98.45 & 83.74 & 96.04 & 96.90 \\
				\multicolumn{1}{c}{\multirow{1}{*}{RGB+D}} & \textbf{100} & \textbf{99.74} & \textbf{100} & \textbf{99.05} & \textbf{87.03} & \textbf{96.96} & \textbf{97.34} \\
				\hline 
		\end{tabular}}
		\label{table_2}
	\end{center}
\vspace{-1.0em}
\end{table}

\textit{Effectiveness of MCFLNet}. To evaluate the effectiveness of MCFLNet, we train a two-stream feature extraction network based on resnet18 using RGB and the depth map generated by FFDGNet. The RGB-D FR accuracy of the testing set was used as the benchmark. We then used the same images to train and test on MCFLNet, and the results are reported in Table \ref{table_3}. It can be observed that MCFLNet achieved a significant accuracy improvement of 4.92\% and 0.82\% on the TM and AVG subsets of Lock3DFace, respectively. Furthermore, we improved the RGB-D FR accuracy on IIIT-D by 2.55\%. This is attributed to the fact that MCFLNet can force the network to learn more complementary features, allowing us to utilize more discriminative information for RGB-D FR.
\begin{table}[t]
	\begin{center}
		\setlength{\belowcaptionskip}{0.4cm}
		\caption{The FR accuracy of RGB and the depth images generated by FFDGNet. (a) two-stream feature extraction network based on ResNet18, (b) the feature extraction network MCFLNet.}
		\setlength{\tabcolsep}{1mm}{
			\begin{tabular}{cccccccc} 
				\hline
				\multicolumn{1}{c}{\multirow{2}{*}{Method}} & \multicolumn{6}{c}{Lock3DFace} & IIIT-D \\ \cline{2-7}
				\multicolumn{1}{c}{\multirow{2}{*}{}} & NU & FE & OC & PS & TM & AVG & AVG \\ \hline 
				\multicolumn{1}{c}{\multirow{1}{*}{(a)}} & 100 & 99.74 & 100 & 99.05 & 87.03 & 96.96 & 97.34 \\
				\multicolumn{1}{c}{\multirow{1}{*}{(b)}} & 100 & 100 & 100 & 98.80 & \textbf{91.95} & \textbf{97.78} & \textbf{99.89} \\
				\hline 
				
		\end{tabular}}
		\label{table_3}
	\end{center}
\vspace{-1.0em}
\end{table}

\subsection{Ablation study}
Firstly, we evaluate the effectiveness of the MFS loss in FFDGNet. We generate depth images using FFDGNet with and without the MFS loss, and then we use RGB and the generated depth images to conduct RGB-D FR experiments on MCFLNet. Table \ref{table_8} shows that the MFS loss improves the AVG accuracy by 0.34\% and 1.33\% on Lock3dFace and IIIT-D, respectively. This confirms that our MFS loss can improve the discrimination of generated depth images by enriching facial details.

Secondly, we evaluate the effectiveness of CIC and CFE loss in MCFLNet. In the first line of Table \ref{table_7}, we use RGB and the generated depth images by FFDGNet to train a two-stream feature extraction network based on resnet18 and adopt the test result as benchmark. The rest result of Table \ref{table_7} shows that the CIC loss leads to a 1.33\% improvement on IIIT-D and the CFE loss leads to 0.5\% and 0.55\% improvements on Lock3DFace and IIIT-D, respectively. These results confirm that our CIC and CFE loss can help the network extract more complementary features, and then learn more discriminative features for RGB-D FR.

\begin{table}[htbp]

	\begin{center}
		\setlength{\belowcaptionskip}{0.4cm}
		\caption{Ablation study of FFDGNet on Lock3DFace and IIIT-D datasets.}
		\setlength{\tabcolsep}{0.6mm}{
			\begin{tabular}{cccccccc } 
				\hline
				\multicolumn{1}{c}{\multirow{2}{*}{Method}} &
				\multicolumn{6}{c}{\multirow{1}{*}{Lock3DFace}} & IIIT-D \\ \cline{2-7} \multicolumn{1}{c}{\multirow{1}{*}{}} & NU & FE & OC & PS & TM & AVG & AVG  \\
				w/o MFS &  99.95 & 100 & \textbf{99.15} & \textbf{99.35} & 89.62 & 97.44 & 98.56 \\
				w/ MFS &  \textbf{100} & \textbf{100} & 99 & 98.60 & \textbf{91.95} & \textbf{97.78} & \textbf{99.89} \\
				\hline 
				
		\end{tabular}}
		\label{table_8}
	\end{center}
\vspace{-0.5em}
\end{table}

\begin{table}[htbp]
	\begin{center}
		\setlength{\belowcaptionskip}{0.4cm}
		\caption{Ablation study of MCFLNet on Lock3dFace and IIIT-D datasets.}
		\setlength{\tabcolsep}{2.5mm}{
			\begin{tabular}{cccc } 
				\hline
				\multicolumn{4}{c}{Rank-1 Identification Average Accuracy} \\ \hline
				CFE Loss & CIC Loss & Lock3DFace & IIIT-D \\ \hline
				--- & --- & 96.96 & 97.34 \\
				--- & \Checkmark & 97.02 & 98.67 \\
				\Checkmark & --- & 97.52 & 99.22 \\
				\Checkmark & \Checkmark & \textbf{97.78} & \textbf{99.89} \\

				\hline 
				
		\end{tabular}}
		\label{table_7}
	\end{center}
\vspace{-2.0em}
\end{table}

\subsection{Comparison with the state-of-the-art methods}
In this section, we evaluate the performance of our proposed method for RGB-D FR using RGB and face depth images generated by FFDGNet, and compare it with several state-of-the-art methods. The obtained results are presented in Table \ref{table_6} and Table \ref{table_5}. As can be observed, our approach achieves superior performance to state-of-the-art methods on both IIIT-D and Lock3DFace datasets. Specifically, our recognition network attains an accuracy of 99.89\% on IIIT-D, which is 0.19\% higher than the state-of-the-art result. Additionally, on Lock3DFace, our proposed method achieves an accuracy of 97.78\%, which is 1.74\% higher than the current state-of-the-art.

\begin{table}[t]
	
	\begin{center}
		\setlength{\belowcaptionskip}{0.5cm}
		\caption{The comparison of rank-1 identification accuracy between our method and state-of-the-art methods on IIIT-D. }
		\setlength{\tabcolsep}{8mm}{
			\begin{tabular}{cc} 
				\hline
				Models &  Average \\ \hline
				Kumar \textsl{et al.} \cite{bl2021rgb} &  97.50 \\
				Goswami \textsl{et al.} \cite{goswami2013rgb}  & 98.10 \\
				Zhang \textsl{et al.} \cite{zhang2018rgb} & 98.60 \\
				Chowdhury \textsl{et al.} \cite{chowdhury2016rgb}  & 98.71 \\
				Uppal \textsl{et al.} \cite{uppal2020attention}  & 99.40 \\
				Uppal \textsl{et al.} \cite{uppal2021two} & 99.40 \\
				Chiu \textsl{et al.} \cite{chiu2021high} & 99.70 \\
				Ours \textsl{et al.} & \textbf{99.89} \\
				\hline 
				
		\end{tabular}}
		\label{table_6}
	\end{center}
\vspace{-2.0em}
\end{table}

\subsection{Parameter analysis}

\begin{figure}[htbp]
	\begin{center}
		\includegraphics[width=0.8\linewidth]{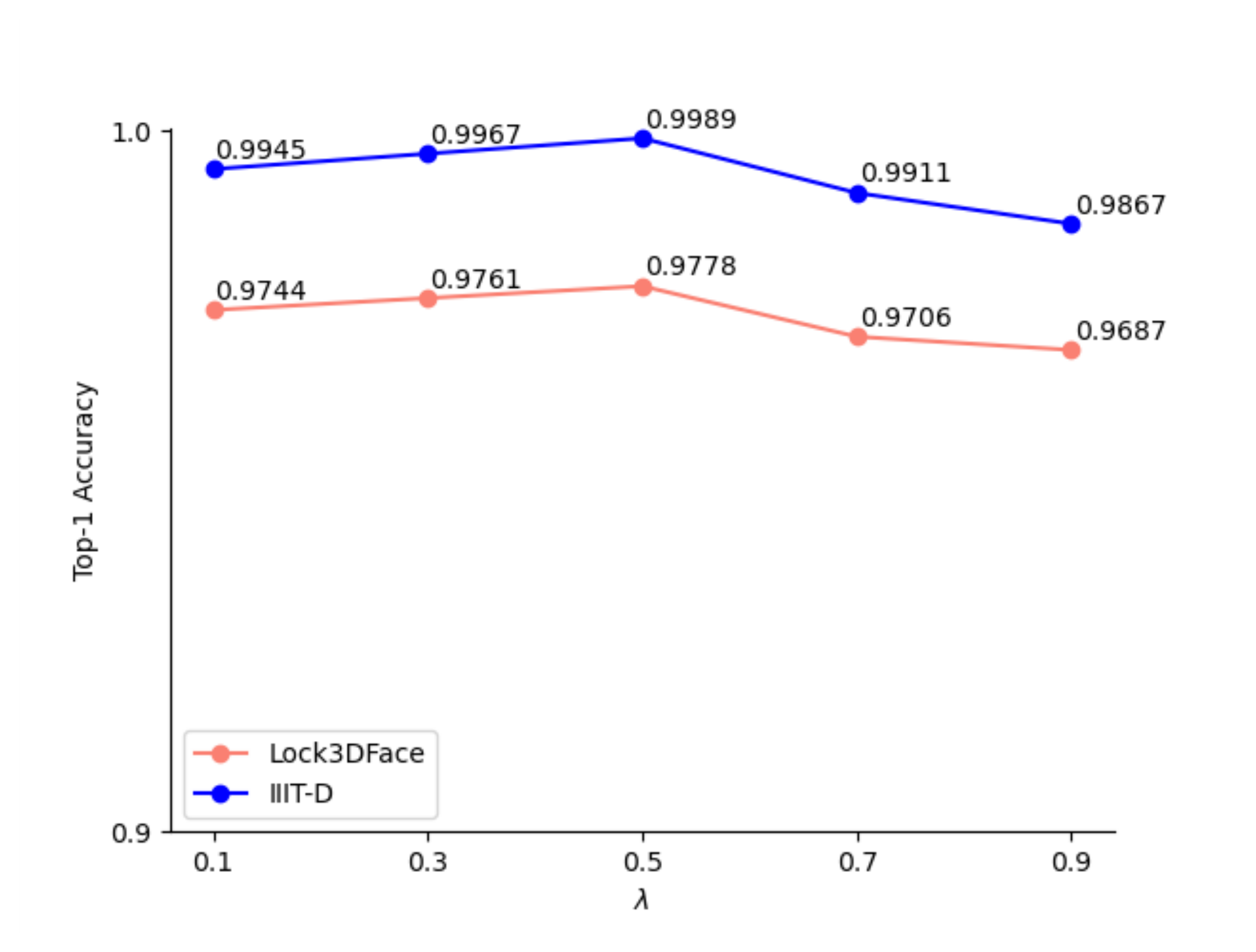}
	\end{center}
	\setlength{\abovecaptionskip}{-0.2cm}\caption{Analysis of rank-1 accuracy versus hyper-parameter for $\lambda$ on two datasets.}
	\label{figure_5}
\vspace{-1.0em}
\end{figure}

In the proposed FR network, we find that the classification loss of the MCFLNet is an order of magnitude higher than the other two items during early training. To balance the gradient of the three loss terms in the backpropagation, we set the weight of the classification loss to a number less than 1. Fig. \ref{figure_5} shows the experimental result with different weight parameters $\lambda$ on two datasets. It can be seen that the optimal weight is 0.5 when $\lambda$ is set to [0,1]. Thus, we set it to 0.5 in our paper. 

\begin{table*}[htbp]
	\begin{center}
		\setlength{\belowcaptionskip}{0.4cm}
		\caption{The rank-1 identification accuracy of our method and state-of-the-art methods on Lock3DFace. }
		\setlength{\tabcolsep}{2mm}{
			\begin{tabular}{cccccccc} 
				\hline
				Models & Inputs & NU & FE & OC & PS & TM & AVG \\ \hline
				Lin \textsl{et al.} \cite{lin2021high} & GT Depth+Gen Depth & 99.95 & 97.31 & 80.97 & 73.61 & 61.67 & 86.55 \\
				Jiang \textsl{et al.} \cite{jiang2021pointface} & Point Cloud+normal & 99.46 & 98.52 & 80.67 & 73.69 & 67 & 87.18 \\
				Mu \textsl{et al.} \cite{mu2021refining} & 3D Face Depth & 99.96 & 96.83 & 64.98 & 69.25 & 73.68 & 82.23 \\ 
				Mu \textsl{et al.} \cite{mu2019led3d} & RGB+GT Depth & 99.62 & 97.62 & 68.93 & 64.81 & 64.97 & 81.02 \\
				Cui \textsl{et al.} \cite{cui2018improving} & RGB+Gen Depth & 100 & 99.59 & 100 & 98.45 & 83.74 & 96.04 \\
				Ours & RGB+Gen Depth & \textbf{100} & \textbf{100} & \textbf{99.60} & \textbf{98.80} & \textbf{90.24} & \textbf{97.78} \\
				\hline 
				
		\end{tabular}}
		\label{table_5}
	\end{center}
	\vspace{-2.0em} 
\end{table*}

%-----------------------------------------conclusion---------------------------------------
\section{Conclusion}
In this paper, we introduce a two-stage FR method that comprises of FFDGNet and MCFLNet. FFDGNet could generate facial depth images with rich facial details, which could provide more additional information to enhance the accuracy of 2D FR in complex scenarios. MCFLNet can improve the performance of multi-modal RGB-D FR by learning more complementary features through constraining the similarity of features in four subspaces of two modalities. Our experimental results demonstrate that incorporating facial contour details into generated depth images and enhancing the learning of modality-specific and modality-shared features could improve the accuracy of RGB-D FR in complex scenarios.

\bibliographystyle{elsarticle-num}
\bibliography{refs}

\end{document}